\documentclass[10pt,twocolumn,letterpaper]{article}

\usepackage[pagenumbers]{cvpr} %

\usepackage[disable]{todonotes}

\usepackage{xcolor}
\definecolor{turquoise}{cmyk}{0.65,0,0.1,0.3}
\definecolor{purple}{rgb}{0.65,0,0.65}
\definecolor{dark_green}{rgb}{0, 0.5, 0}
\definecolor{orange}{rgb}{0.8, 0.6, 0.2}
\definecolor{red}{rgb}{0.8, 0.2, 0.2}
\definecolor{darkred}{rgb}{0.6, 0.1, 0.05}
\definecolor{blueish}{rgb}{0.0, 0.3, .6}
\definecolor{lightgrey}{rgb}{0.9, 0.9, .9}
\definecolor{lightred}{rgb}{0.9, 0.7, .7}

\definecolor{pink}{rgb}{1, 0, 1}
\definecolor{greyblue}{rgb}{0.25, 0.25, 1}

\definecolor{thirdbestcolor}{rgb}{1,1, 0.6}
\definecolor{secondbestcolor}{rgb}{1, 0.9, 0.6}
\definecolor{firstbestcolor}{rgb}{1, 0.6, 0.6}

\usepackage{lipsum}  
\usepackage{nicefrac}
\usepackage{multirow} 
\usepackage{pifont}

\usepackage{times}
\usepackage{epsfig}
\usepackage{amsmath}
\usepackage{amssymb}
\usepackage{booktabs}
\usepackage{makecell}
\usepackage{tabularx}
\usepackage{bbm}
\usepackage{caption} 
\usepackage{subcaption}
\usepackage{animate}

\usepackage{xspace}
\usepackage{graphicx}

\newcommand{\radiancefield}{radiance field\xspace}

\usepackage[
final,tracking=true,kerning=true,spacing=true,factor=1100,stretch=10,shrink=10]{microtype}
\microtypesetup{protrusion=false}

\usepackage{tcolorbox}
\AtBeginEnvironment{tcolorbox}{\small}

\usepackage[accsupp]{axessibility}

\definecolor{cvprblue}{rgb}{0.21,0.49,0.74}
\usepackage[pagebackref,breaklinks,colorlinks,allcolors=cvprblue]{hyperref}
\usepackage{verbatim}

\usepackage[capitalize]{cleveref}
\crefname{section}{Sec.}{Secs.}
\Crefname{section}{Section}{Sections}
\Crefname{table}{Table}{Tables}
\crefname{table}{Tab.}{Tabs.}

\def\paperID{10340} %
\def\confName{CVPR}
\def\confYear{2025}

\newcommand{\ours}{RelationField\xspace}
\newcommand{\mytitle}{\ours: Relate Anything in Radiance Fields}

\newcommand{\bc}{\mathbf{c}}
\newcommand{\bd}{\mathbf{d}}

\newcommand{\bi}{\mathbf{i}}

\newcommand{\bo}{\mathbf{o}}
\newcommand{\bp}{\mathbf{p}}

\newcommand{\br}{\mathbf{r}}
\newcommand{\bs}{\mathbf{s}}

\newcommand{\bx}{\mathbf{x}}

\newcommand{\bz}{\mathbf{z}}

\newcommand{\nR}{\mathbb{R}}
\newcommand{\nS}{\mathbb{S}}

\newcommand{\cI}{\mathcal{I}}

\newcommand{\cL}{\mathcal{L}}

\newcommand{\cP}{\mathcal{P}}

\newcommand{\cR}{\mathcal{R}}
\newcommand{\cS}{\mathcal{S}}

\makeatletter
\DeclareRobustCommand\onedot{\futurelet\@let@token\@onedot}
\def\@onedot{\ifx\@let@token.\else.\null\fi\xspace}

\def\etal{et~al\onedot}

\makeatother

\newcommand{\boldparagraph}[1]{\vspace{0.5em}\noindent{\bf #1.}}

\renewcommand{\paragraph}[1]{\boldparagraph{#1}}

\definecolor{darkgreen}{rgb}{0,0.7,0}
\definecolor{newyellow}{rgb}{1,0.8,0.05}
\definecolor{newgreen}{rgb}{0.2,0.8,0.2}

\title{\mytitle}

\author{
Sebastian Koch$^{1,2}$\qquad Johanna Wald$^{3}$\qquad Mirco Colosi$^2$\qquad Narunas Vaskevicius$^{2}$ \\Pedro Hermosilla$^4$\qquad Federico Tombari$^{3,5}$\qquad Timo Ropinski$^1$ \vspace{0.3cm}\\ 
{ $^1$\text{University Ulm} \quad $^2$\text{Bosch Center for AI} \quad $^3$\text{Google} \quad $^4$ \text{TU Vienna} \quad  $^5$\text{TU Munich}}
}

\begin{document}
\twocolumn[{%
\renewcommand\twocolumn[1][]{#1}%
\maketitle
\vspace{-3.0em}
\begin{center}
    \includegraphics[width=\linewidth]{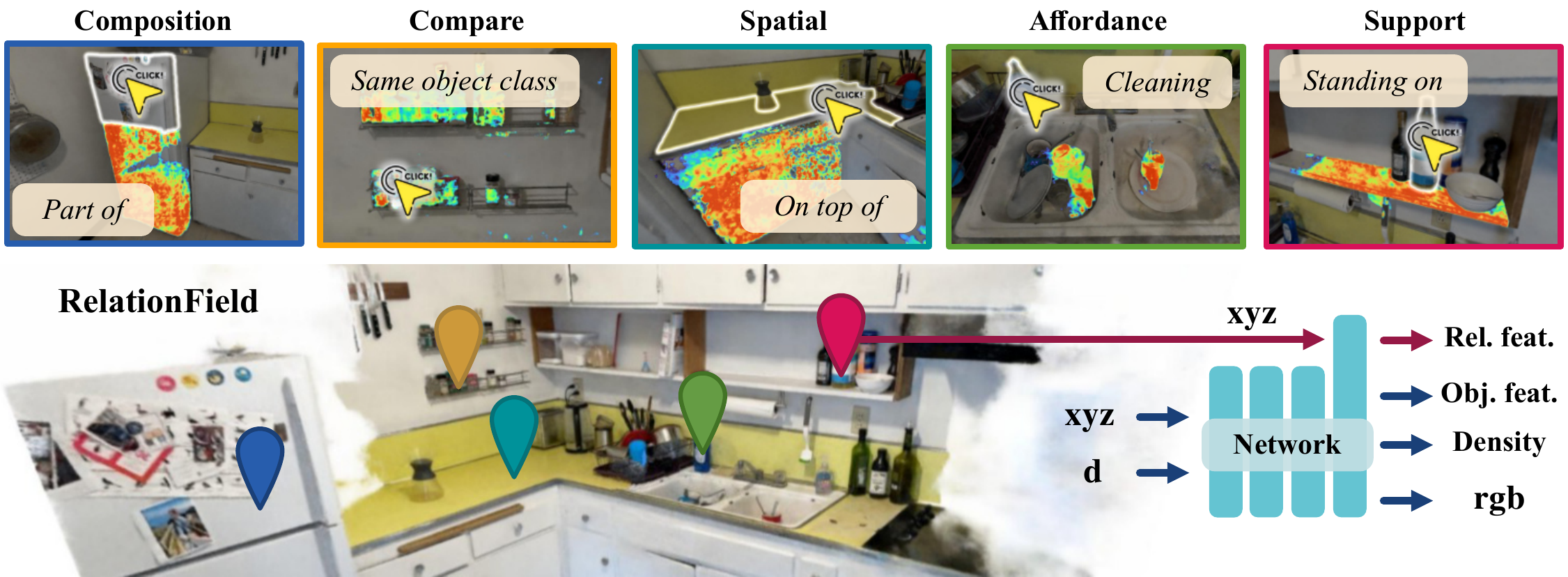}
    \captionof{figure}{\textbf{Open-Vocabulary Relationship Understanding.} We propose \ours, 
     the first framework to extract open-vocabulary inter-object relationships directly from neural radiance fields.
     \ours can answer a wide variety of relationship queries, such as ``composition", ``compare", ``spatial", ``affordance" and ``support" relationships.
     }
\label{fig:teaser}
\end{center}
}]

\begin{abstract}
\vspace*{-0.7cm}

Neural radiance fields are an emerging 3D scene representation and recently even been extended to learn features for scene understanding by distilling open-vocabulary features from vision-language models. However, current method primarily focus on object-centric representations, supporting object segmentation or detection, while understanding semantic relationships between objects remains largely unexplored. To address this gap, we propose \ours, the first method to extract inter-object relationships directly from neural radiance fields. \ours represents relationships between objects as pairs of rays within a neural radiance field, effectively extending its formulation to include implicit relationship queries. To teach \ours complex, open-vocabulary relationships, relationship knowledge is distilled from multi-modal LLMs.
To evaluate \ours, we solve open-vocabulary 3D scene graph generation tasks and relationship-guided instance segmentation, achieving state-of-the-art performance in both tasks.
See the project website at \href{https://relationfield.github.io}{relationfield.github.io}.
\end{abstract}

\vspace*{-0.4cm}

\section{Introduction}
\label{sec:intro}
\vspace*{-0.15cm}
3D scene understanding bridges the gap between the 
physi-
\filbreak\noindent
cal 
and the digital world, by enabling machines to perceive 
environments in a way similar to humans. 
In robotics, 3D scene understanding is required to navigate complex environments, interact with objects, and perform tasks autonomously. In AR/VR it enables realistic and immersive experiences, e.g., by allowing accurate placing of and interacting with virtual content in the real world.
Notably, many applications require a level of understanding that goes beyond just localizing and segmenting a known list of objects categories~\cite{Schult23, misra2021-3detr, qi2019deep, choy20194d} but are also able to segment novel entities beyond the closed-set class assumption~\cite{Peng_2023_CVPR, Ha_2022_CORL, Takmaz_2023_NeurIPS}.

True holistic and adaptable scene understanding needs to go a step further and not only reconstruct and identify individual objects within a scene but also understand complex inter-object relationships, functionalities, and the overall context of the environment. This aspect of scene understanding, particularly the ability to recognize and reason about relationships between objects, is often overlooked. Yet, it is essential to interact with the surroundings in a sophisticated, adaptive and natural manner. 
Significant progress has been made in understanding relationships in 2D images, mainly driven by the exploration of foundation models~\cite{pmlr-v139-radford21a,caron2021emerging,Kirillov_2023_ICCV} and in particular by multi-modal LLMs~\cite{achiam2023gpt,dubey2024llama}. These models are extremely powerful, although they primarily operate 
on 2D representations and do not fully leverage the richness of 3D data. 

3D scenes provide more complete captures of the environment and are able to represent a high level of complexity, with overlapping objects and occlusions that make it difficult to consistently infer relationships with 2D models alone. 3D approaches have been shown to reduce per-frame noise and resolve occlusions. Despite this advantage, 3D foundation models have yet to emerge, as the data available in 3D remains limited compared to 2D.

3D scene graphs on the other hand, are a promising and compact representation for scene understanding and capture not only scene objects but also inter-object relationships. However, several scene graph approaches either rely on a closed set of relationships~\cite{Wald_2020_CVPR,wang2023vl,Koch_2024_WACV,Wu_2021_CVPR}, depend on class-agnostic instance segmentation~\cite{Koch_2024_CVPR}, and/or require an explicit 3D representation such as point clouds.

A recent work, Open3DSG~\cite{Koch_2024_CVPR}, distills relationship knowledge from foundation models~\cite{pmlr-v139-radford21a,instructblip} into a 3D graph neural network, which can then predict open-vocabulary graphs.  
Capturing both objects and relationships with open-vocabulary features allows capturing a wide range of objects, functions, and relationships without prior training on specific object or relationship classes. This flexibility is crucial for handling the diversity and complexity of real-world scenes. However, Open3DSG still relies on given class-agnostic instance segmentation~\cite{Koch_2024_CVPR} and is bound by the quality of the explicit 3D mesh representation of the underlying dataset. These approaches furthermore require the availability of depth sensors.  
In contrast to 3D scene graphs, radiance fields are 3D representations that do not require 3D sensor data, but instead represent 3D scenes solely based on a set of posed 2D images~\cite{Mildenhall_2020_ECCV, Kerbl_2023_SIGGRAPH}. While they were first introduced for novel view synthesis and 3D reconstruction, they have since then been extended in several works to also capture semantic information~\cite{Siddiqui_2023_CVPR, Kerr_2023_ICCV, Engelmann_2024_ICLR, Qin_2024_CVPR}. 

LERF~\cite{Kerr_2023_ICCV}, as well as a few follow-up works~\cite{Engelmann_2024_ICLR,Kim_2024_CVPR,Qin_2024_CVPR} present alternative approaches to distill features from 2D foundation models, such as CLIP~\cite{pmlr-v139-radford21a}, DINO~\cite{caron2021emerging} or SAM \cite{Kirillov_2023_ICCV}, into 3D by means of radiance fields. Yet, these approaches predominantly focus on object-centric semantic features, limiting their application in high-level scene reasoning tasks.

To enable holistic and high-level scene reasoning tasks based on neural radiance fields, we propose \ours, a rich radiance field representation that learns open-vocabulary features for objects and their relationships. %
This allows us to reason about complex scenes and object interactions such as compositional, spatial, support, or affordances, see Fig.~\ref{fig:teaser}. In summary, this work has the following contributions:\looseness=-1
\begin{itemize}
\item We present the first method for open-vocabulary scene segmentation enabling interactive and textual relationship queries by extending the semantic neural radiance formulation with inter-object relationships distilled from a foundation model into a dense and multi-view consistent 3D representation.
\item %
This novel representation not only facilitates relationship-based queries but also allows us to obtain state-of-the-art 3D scene graphs -- making it the first time scene graphs have been inferred from neural radiance fields.
\item Furthermore, we introduce a new task -- relationship-guided instance segmentation -- on ScanNet++~\cite{Yeshwanth_2023_ICCV}. This task involves segmenting an instance based on an object-relationship search query, e.g., ``picture \textit{standing on} the shelf", providing a benchmarking for future research in this direction.
\end{itemize}

\begin{figure*}
    \centering
    \includegraphics[width=\linewidth]{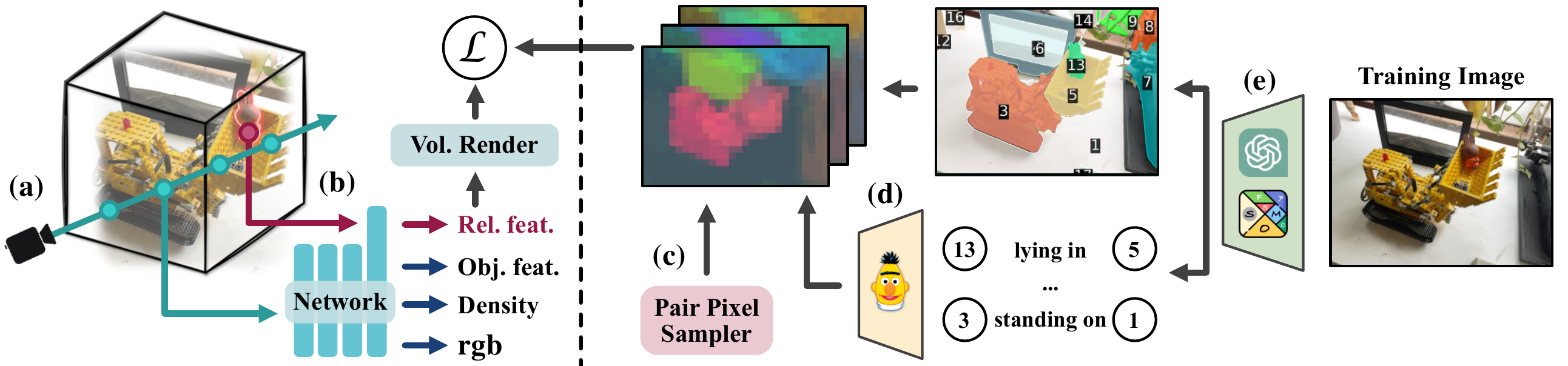}
    \caption{\textbf{\ours Training.} \textit{Left:} \ours learns a 3D feature field (a) that can be queried with a relationship query location (b) which changes the relationship field of the 3D volume depending on what position is selected. The relationship feature is sampled and rendered along a ray according to NeRF's rendering weights. The language loss maximizes the cosine similarity between the extracted sparse features from the 2D views and the rendered 3D relationship features. \textit{Right:} We estimate 2D relationship proposals from a multi-model LLM prompted with SoM (e) for each training view and encode extracted textual relationship description into the image plane (d). A pair pixel sampler samples subject and object pixels (c) for which the relationship feature is distilled into the 3D volume.}
    \label{fig:method}
\end{figure*}

\section{Related Work} \label{sec:related_work}

\vspace*{-10pt}

\boldparagraph{Open-Vocabulary 3D Scene Understanding} 
Recent 3D scene understanding approaches for detection, semantic segmentation, or instance segmentation have moved from closed-set categories~\cite{Schult23, misra2021-3detr, qi2019deep, choy20194d}
to open-vocabulary, removing the limitation to a pre-defined vocabulary. To do so, 2D features from vision-language models (VLMs) are lifted into 3D by either using feature distillation and feature lifting. The latter extract vision-language features directly on 2D images and then project these to 3D by utilizing depth or by separately training 2D and 3D feature encoders that are combined at inference time~\cite{Ha_2022_CORL, Takmaz_2023_NeurIPS, Nguyen_2024_CVPR, Delitzas_2024_CVPR, Huang_2024_ECCV}. Feature distillation on the other hand, trains a 3D model using semantic features extracted from a VLM from posed 2D images~\cite{Peng_2023_CVPR, Koch_2024_CVPR, Engelmann_2024_ICLR} and does not assume the availability of 2D frames at test time. Both feature lifting and distillation methods require 2D and 3D data either for training or for inference.\looseness=-1

While open-vocabulary 3D scene understanding approaches have shown impressive progress in semantic object segmentation, they do not holistically capture the scene lacking knowledge about high-level compositions and/or inter-object relationships.

\boldparagraph{Relationships in 3D Scenes}
Understanding the full 3D scene involves extracting compositional knowledge and relationships between objects and has been shown to improve object-centric predictions~\cite{Wu_2021_CVPR, Kulkarni_2019_ICCV}. 3D scene graphs~\cite{Armeni_2019_ICCV,Wald_2020_CVPR} have emerged as the predominant representation for modeling these relationships with applications in several different tasks such as place recognition~\cite{Wald_2020_CVPR}, registration~\cite{Sarkar_2023_ICCV}, change detection~\cite{Wald_2020_CVPR,looper22vsg}, task planning~\cite{Agia_2022_PMLR, rana2023sayplan, liu2024delta}, and navigation~\cite{Werby_2024_RSS}. By representing objects as nodes into graphs and explicitly encoding their connections (spatial, semantic, etc.) as edges, 3D scene graphs offer an efficient representation of the environment. \cite{Armeni_2019_ICCV} proposes to represent buildings, rooms, objects, and cameras as 3D scene graphs and later works extended this idea by learning hierarchical 3D scene graphs directly from sensor data~\cite{Rosinol_2020_RSS,hughes2022hydra,Rosinol_2021_SAGE}. On the other hand,~\cite{Wald_2020_CVPR} introduced semantic 3D scene graphs, focusing more on the semantic components of a scene including inter-object relationships. Subsequent works have advanced this research area by refining semantic 3D scene graphs from point clouds using scene priors \cite{Zhang_2021_Neurips}, pre-training~\cite{Koch_2024_WACV, koch_2024_3DV} and improved message passing in graphs~\cite{Wu_2021_CVPR, Wu_2023_CVPR}.

While all these works have a close-set assumption, only a few very recent works have investigated the use of VLMs and large language models (LLMs) to obtain open-vocabulary scene graphs which capture a more flexible representation of the environment~\cite{Koch_2024_CVPR, Maggio2024Clio, Chen_2024_CVPR, gu2024conceptgraphs, cheng2024spatialrgpt}. However, these approaches often require depth data and a complete and explicit 3D representation of the scene e.g. in the form of a 3D mesh or point cloud~\cite{Wald_2020_CVPR, Koch_2024_WACV, Koch_2024_CVPR} which often is not available or of poor quality. 

\boldparagraph{Radiance and Feature Fields}
Radiance Fields~\cite{Mildenhall_2020_ECCV, Kerbl_2023_SIGGRAPH} were first introduced for novel view synthesis and have the benefit that they do not require explicit 3D supervision. Recently, radiance fields have been adapted for several different 3D scene understanding tasks such as segmentation~\cite{Zhi_ICCV_2021, Siddiqui_2023_CVPR} or detection \cite{Xu_2023_ICCV, Hu_2023_CVPR}.
Notably, some methods propose to extend radiance fields to predict features obtained from 2D foundation models in 3D. For instance, LERF~\cite{Kerr_2023_ICCV} and OpenNeRF~\cite{Engelmann_2024_ICLR} learn vision-language features using a separate MLP-head in the NeRF model to produce CLIP~\cite{pmlr-v139-radford21a} embeddings for open-vocabulary 3D segmentation. %
Similarly, GARField~\cite{Kim_2024_CVPR} learns instance embeddings using a contrastive formulation provided by SAM~\cite{Kirillov_2023_ICCV} using a separate MLP-head in their NeRF. Among others, LangSplat~\cite{Qin_2024_CVPR} and ClickGaussians~\cite{Choi_2024_arxiv} extend these ideas to Gaussian Splatting for faster training and rendering. While these works show impressive results, they mainly investigate object-centric semantics and also do not explore the composition of a scene or object relationships.

Inspired by these works, our method learns open-vocabu\-lary vision-language features directly from multiple posed 2D views. Therefore, we similarly do not require any explicit 3D scene representation in the form of depth or point cloud data. Instead, our approach aims to obtain open-vocabulary scene understanding beyond objects by also encoding object relationships, creating a consistent and rich representation. This way, our approach -- as the first of its kind -- supports interactive relationship queries and allows to extract 3D semantic scene graphs directly from the radiance field.

\section{Method}
\label{sec:method}
Given a set of posed RGB images, our goal is to 
build a queryable 
3D representation of the scene that supports understanding object instances using open-vocabulary object and relationship descriptions.
To achieve this, we introduce a novel approach, \ours, as illustrated in~\cref{fig:method}. 
Our proposed approach is independent of the underlying radiance field, and can be adapted to NeRFs \cite{Mildenhall_2020_ECCV} as well as Gaussian Splatting \cite{Kerbl_2023_SIGGRAPH}, in the following section we demonstrate how our method
incorporates implicit open-set relationship feature prediction into NeRFs\footnote{An adaption to Gaussian Splatting is detailed in the supplementary material.} \cite{Mildenhall_2020_ECCV}, enabling the querying of arbitrary object and relationship concepts within a continuous volumetric 3D scene representation. To enhance NeRF with object-centric semantics, we distill CLIP-feature~\cite{pmlr-v139-radford21a} prediction and SAM~\cite{Kirillov_2023_ICCV} supervision for instance grouping of each ray. %
Our method is the first to introduce an implicit open-set relationship feature prediction head as explained in~\cref{sec:relastionship_field}. It is supervised by the embedded features of a multi-modal LLM using set-of-mark prompting (SoM)~\cite{Yang_2023_setofmark} (see \cref{sec:2d_supervision}). The learned \ours then can be queried to retrieve relationships such as ``the light switch \textit{turns on} the lamp" by defining the predicate ``turns on" as a pair of input rays within the feature field for all rays that hit the \textit{light switch} and \textit{lamp} (see \cref{sec:relationship_query}).

\subsection{\ours}
\label{sec:relastionship_field}
\boldparagraph{Radiance Field}
A radiance field describes a function that %
models the color $\bc \in [0,1]^3$ and density $\sigma \in [0,\infty)$ for a given 3D point $\bx \in \nR^3$ and ray direction $\bd \in \nS^2$. Mildenhall \etal \cite{Mildenhall_2020_ECCV} first proposed to model this implicit function 
as a neural radiance field (NeRF) that implements a multilayer perceptron $f$ with the training objective of learning the parameters $\theta$ with supervision from multi-images of the scene
\begin{equation}
    f_\theta(\bx, \bd)  \mapsto (\bc, \sigma).
\end{equation}

\boldparagraph{Object-level Semantics in Radiance Fields}
To learn object-level open-vocabulary instances within the radiance field, we extend NeRF with two additional output embedding heads: one predicts open-vocabulary features $\bs$ in the CLIP embedding space, inspired by \cite{Engelmann_2024_ICLR}, and the other predicts a grouping embedding $\bi$ that co-locates rays of the same instance in the same region of the embedding space for easy instance clustering,
similar to \cite{Kim_2024_CVPR}.
The open-vocabulary feature is therefore defined as a tuple $\bo=(\bs,\bi)$ of semantic and instance features.
These object-level open-vocabulary features allow us to query object entities but do not capture relationships.
Therefore, it is necessary to model relationships explicitly.

\boldparagraph{Relationship Semantics in Radiance Fields}
Unlike radiance fields, which only predict color and density for a point $\bx$, relationship modeling requires an additional point $\bz$ to specify the relationship between $\bx$ and $\bz$.
Therefore, to capture relationships within the radiance field, we extend the input by an additional implicit query location $\bz \in \nR^3$ (\cref{fig:method}b). With this query location, our approach implicitly models the relationship feature $\br$ between the ray $(\bx,\bd)$ 
and the location $\bz$. 
The relationship feature $\br$ is located within the language embedding space and can be queried for arbitrary relationships based on the cosine similarity. 

The complete function $g_\theta$ that models the color, density, open-vocabulary instance feature as well as open-vocabulary relationships of the objects in the 3D scene is given by
\begin{equation}
    g_\theta(\bx, \bd, \bz)  \mapsto (\bc, \sigma, \bo, \br).
\end{equation}

\subsection{Relationship Supervision}
\label{sec:2d_supervision}
While vision-language models such as CLIP \cite{pmlr-v139-radford21a} excel at modeling individual objects and concepts, 
their understanding of relationships remains limited \cite{yuksekgonul2022and}.
To address this, we distill relationship knowledge from multi-modal LLMs, which better represent complex relationships.
However, a challenge arises because multi-modal LLMs produce textual descriptions, while models like CLIP generate pixel- or patch-level features that can be queried using various text encodings. 
Our goal is to transfer relationship features into the \radiancefield representation, enabling open-vocabulary querying similar to object-centric approaches with CLIP \cite{Engelmann_2024_ICLR,Kerr_2023_ICCV}.
The following paragraphs outline our approach for extracting such high-dimensional, pixel-aligned features from multi-modal LLMs, effectively bridging the gap between textual understanding and visual feature extraction.

\boldparagraph{Set-of-Mark (SoM)}
To extract dense pixel-aligned visual relationship features, we utilize SoM prompting \cite{yang2023setofmark}. SoM is a visual prompting approach that enhances the visual grounding abilities of multi-modal LLMs by overlaying marks, masks, or bounding boxes to help the model answer fine-grained visual questions. By using SoM over a direct approach, it has been shown, that it improves the spatial reasoning of LLMs, such as GPT-4~\cite{yang2023setofmark}.

\boldparagraph{Feature extraction}
To generate sparse high-dimensional pixel-aligned visual relationship features, 
we use SAM~\cite{Kirillov_2023_ICCV} to extract $m$ segmentation masks each corresponding to a detected object in the image from a training view.
Using these masks, we annotate the image with alphanumeric marks for each segmented object following the SoM prompting technique.
Next, we prompt a multi-modal LLM to identify and extract inter-object relationships for closely positioned marked object pairs (\cref{fig:method}e)\footnote{A detailed analysis of our prompting technique is provided in the supplementary material.}. 
The output text $t$ includes a textual description of the relationships between object pairs $(i, j)$ using the identifiers from the SoM annotations.
Each textual relationship description $t_{ij}$ is then encoded to a high-dimensional feature representation $\phi_{t_{ij}}$ using an encoder-only language model such as \cite{kenton2019bert}, resulting in $d$ dimensional features for each relationship (\cref{fig:method}d). 
These features are then projected onto the image plane using the SAM segmentation masks and the SoM marks as a reference to generate a 
high-dimensional feature representation of the extracted relationships that are aligned with the pixel locations of the objects in the image.

\boldparagraph{Training}
During training we randomly sample ray and query origins uniformly throughout the input views in a pairwise manner using a pair-pixel sampler (\cref{fig:method}c). 
Using the density prediction of the \radiancefield, we estimate the query positions along the ray of the query origin. Ray and query samples are concatenated and fed together into an MLP-head that predicts the relationship feature along the sample ray.
The feature is rendered onto the image plane using the \radiancefield's rendering weights. We minimize a loss 
\begin{equation}    
    \cL = 1 - \frac{\br}{||\br||_2} \cdot \frac{\hat{\br}}{||\hat{\br}||_2},
\end{equation}
that maximizes the cosine similarity between the rendered relationship feature $\br$ and the ground-truth relationship feature $\hat{\br}$.
Similarly, the rendered object-centric features, such as color and open-vocabulary semantics, as well as instance features for each ray, are supervised by their respective ray origin features.

\begin{figure*}
    \centering
    \includegraphics[width=\linewidth]{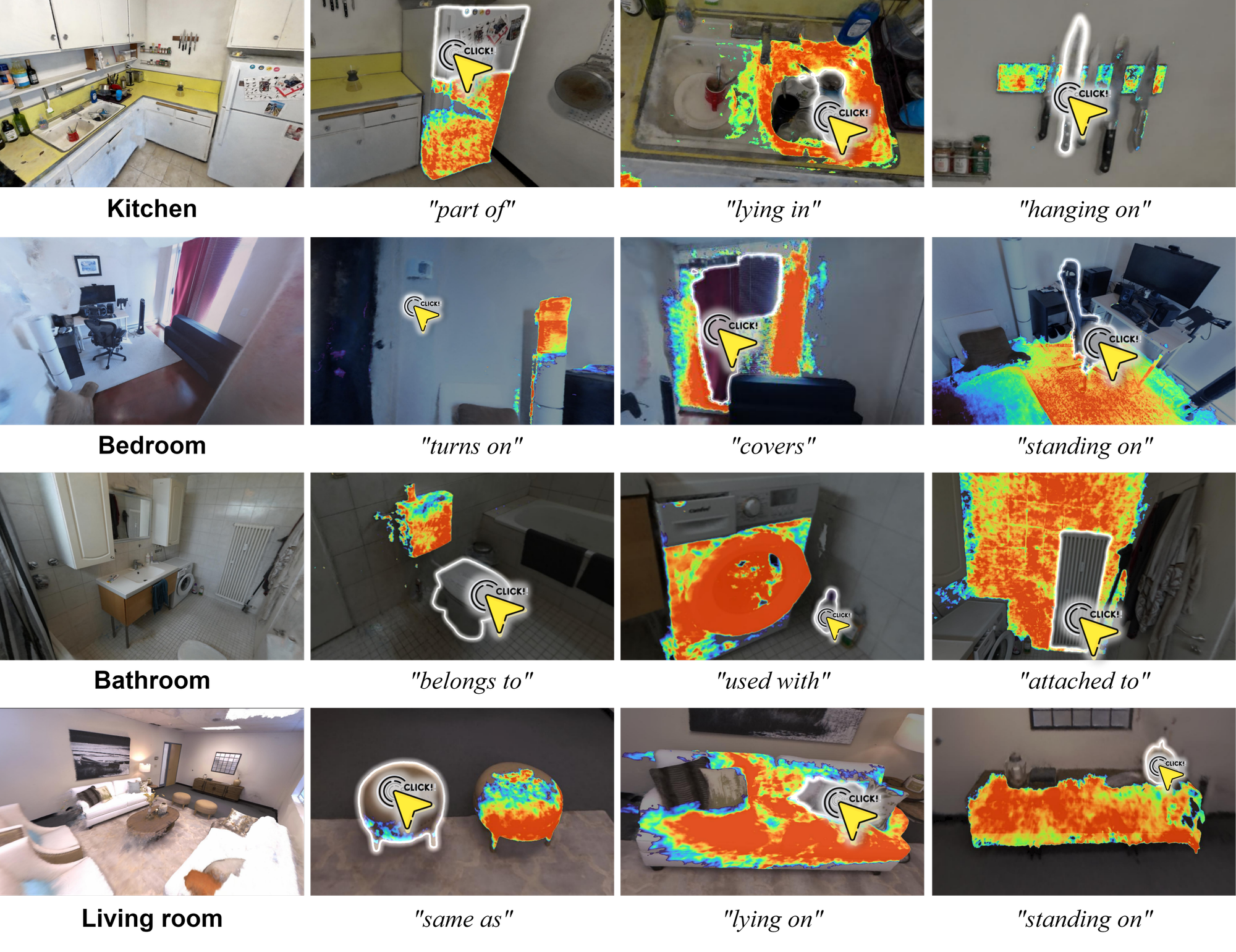}
    \vspace*{-20pt}
    \caption{\textbf{Results with \ours in 4 \textit{in-the-wild} scenes.} Each image shows a rendering from \ours, along with the relationship response for each query relationship. The relevancy score describes the answer of the model to the question: 
    What is \includegraphics[scale=0.02]{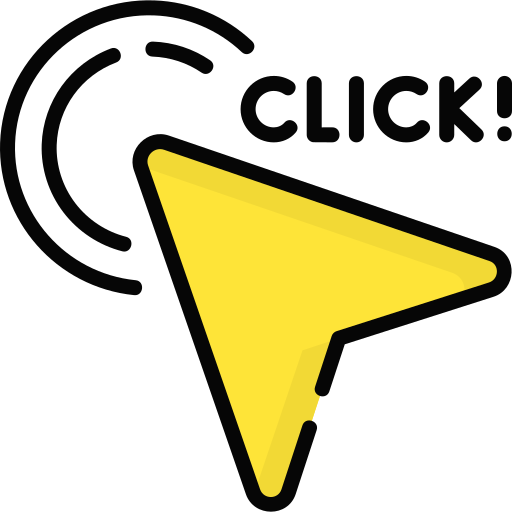} \textit{standing on/attached to/similar to etc.?} 
    For demonstration purposes, we highlight the click as well as the outline of the clicked object, which is not needed when querying the model. Our model is able to understand complex relationships, such as the functionality of light switches or uncommon support structures, such as ``knives hanging on a magnetic mount".
    }
    \label{fig:qualitative}
\end{figure*}

\subsection{Querying \ours}
\label{sec:relationship_query}
To effectively explore and understand the relationships between objects in a scene, it is natural to first identify and query the objects themselves before investigating their inter-object relationships. In this context, \ours supports both object querying and subsequent relationship querying, providing a comprehensive framework for scene understanding.
The querying process of \ours consists of two steps. First, \textit{selecting a query location}, which involves determining for which object in the scene to investigate relationships. This location can be specified directly by the user or chosen based on detected object instances. 
The second step requires to \textit{query a textual relationship} for the selected object. Once a query location is chosen, users can specify a particular relationship they wish to investigate, such as ``standing on" or ``similar to" using a text query. Alternatively, a set of possible relationships for exploration can be provided, which is particularly useful for an open-ended investigation of the scene.

To evaluate the response of a queried relationship, we assign a score to each ray in the \radiancefield by calculating the cosine similarity between the language encoding of the query $\phi_q$, and the relationship embedding, $\br$. However, since it is difficult to interpret the cosine similarity directly without context, we follow the approach introduced by \cite{Kerr_2023_ICCV} and output the pairwise softmax with regard to canonical phrase embeddings $\phi_{canon}$ such as ``and", ``next to" and ``none". The relationship response is then
\begin{equation}
    \rho = \min_i\frac{\exp(\phi_q \cdot \br)}{\exp(\phi^i_{canon}\cdot\br) + \exp(\phi_q \cdot \br)}.
\end{equation}
Intuitively, this softmax probability represents how much the model favors a certain relationship query over no relationship.

\subsection{Implementation Details}
\label{sec:implementation_details}

\ours is built in Nerfstudio \cite{tancik2023nerfstudio} on top of the Nerfacto model for color and density estimation of a given ray from posed training images with known intrinsic and optionally depth supervision. 
We define separate heads to estimate the open-vocabulary semantic object, instance, and relationship feature fields. 
The open-vocabulary segmentation head outputs 768-dimensional features in CLIP~\cite{pmlr-v139-radford21a} / OpenSeg~\cite{OpenSeg_2022_ECCV} embedding space for a given location vector without view-direction. Similarly, the instance head outputs a 256-dimensional grouping feature in the instance embedding space for a given location vector. Our relationship field encodes a pair of location vectors for the ray and query locations by concatenating them and outputs a language-aligned relationship feature of 512 dimensions in the jina-embeddings-v3 \cite{jina-v3} embedding space.
For relationship feature supervision, we use \mbox{GPT-4o}~\cite{achiam2023gpt} to extract relationship features from the training image together with SoM~\cite{Yang_2023_setofmark} using numeric marks and semi-transparent masks. The language outputs are encoded using jina-embeddings-v3~\cite{jina-v3}.

\section{Experiments} \label{sec:experiments}
In the following, we present both qualitative and quantitative results that highlight the capabilities of our method.
To highlight the performance of our method in an \textit{in-the-wild} setting, we provide a qualitative analysis of various relationship queries in different indoor environments in \cref{sec:qualitative_results}.
To quantify \ours performance, we leverage the task of 3D scene graph prediction in \cref{sec:scene_graph_eval}. Our approach outperforms several competitive baselines and establishes a new state-of-the-art on the 3DSSG benchmark.
We then perform comprehensive ablation studies to demonstrate the importance of 3D consistency and knowledge distillation. Specifically, we compare our method against various 2D multi-modal LLMs. Further ablation studies justify our choice of relationship encoders by comparing different multi-modal LLMs for this purpose.
Furthermore, we demonstrate the capabilities of our model in \cref{sec:relationship_inst_refer} by reporting its performance on a new task  -- \textit{relationship-guided 3D instance segmentation} -- which leverages natural language prompts e.g., ``picture standing on the shelf" for 3D segmentation. Notably, our method outperforms all recent open-vocabulary feature fields, demonstrating its ability to understand object relationships accurately.

\subsection{Relationship Segmentation} \label{sec:qualitative_results}
\cref{fig:qualitative}, shows our method's ability to segment relationships. We visualize the model's response for a given textual relationship prompt together with the selected target location. Results are reported on 4 different scenes taken from three datasets: LERF \cite{Kerr_2023_ICCV}, Scannet++ \cite{Yeshwanth_2023_ICCV}, and Replica \cite{straub2019replica}. The scenes consist of several complex object interactions such as compositional relationships like ``the freezer being \textit{part of} the refrigerator", support relationships such as ``the pillow \textit{lying on} the couch", comparative or similarity relationships like ``one ottoman being the \textit{same as} another ottoman", or even affordances such as ``the light switch \textit{turns on} the lamp". The colormap which shows the top 50\% confidence for each query respectively, shows that our model is able to segment these complex relationships.

\subsection{3D Scene Graph Prediction} \label{sec:scene_graph_eval}
Our method's ability to estimate both open-vocabulary relationships as well as object instances enables the generation of 3D scene graphs. The following section details the extraction process of these graphs from our radiance field representation and presents quantitative comparisons against state-of-the-art open-vocabulary 3D scene graph prediction models.
Our proposed approach is not only able to predict open-vocabulary relationships but also open-vocabulary object instances. Combining both predictions enables the inference of open-vocabulary 3D scene graphs.

\boldparagraph{3D Scene Graph Construction}
\todo[inline]{move some stuff to the supplementary or reference for further instructions}

To extract explicit 3D scene graphs from our implicit representation requires an automated querying process. For a fair comparison with point cloud-based methods, we query the radiance field directly on the provided 3D point cloud. This ensures alignment between the extracted graph and the provided point cloud. 
Please note that while our method is trained solely on RGB data, the 3D point cloud is utilized exclusively for evaluation.

To do so, for each 3D point $\bp$ in the point cloud $\cP$, we extract semantic and instance features by querying the \radiancefield at the given location. Since this process is viewpoint-independent, it does not require a ray direction $\bd$. We then identify instances by clustering the instance embeddings using DBSCAN~\cite{9356727}. For each instance $i \in \cI$, the open-vocabulary object embedding $\cS_i$ is obtained by aggregating the respective semantic features.

To extract relationships, each instance $i$, comprising of points $\cP_i$, serves as a query for the relationship field, which predicts relationship embeddings $\cR$ for the remaining point cloud.  The relationship embedding $R_{ij}$ is then obtained for each pair $(i,j)$ by aggregating the relationship embeddings $R_i$ for all other instances $j \in \cI, j \neq i$.

Since the scene graph benchmark evaluates on a closed-set of object and relationship classes, we query with predefined benchmark labels. Object and relationship classes are encoded with CLIP~\cite{pmlr-v139-radford21a} and Jina~\cite{jina-v3} respectively. We then compute the pair-wise cosine similarity between the ground truth label encodings with the predicted embeddings. To evaluate the predictions, we use the top-k recall metric, selecting the top-k highest-scoring classes as introduced in~\cite{lu2016visual}. For relationship prediction, we follow \cite{yang2018graph}; ranking our relationship predictions by multiplying the object and relationship scores. 

Implementation details on the 3D scene graph extraction can be found in the supplementary.

\begin{table}[t]
\tabcolsep=1.75mm
\centering
\scriptsize
\begin{tabular}{@{}lcccccc@{}}
\toprule
& \multicolumn{2}{c}{Object} & \multicolumn{2}{c}{Predicate} & \multicolumn{2}{c}{Relationship} \\
\cmidrule(lr){2-3}\cmidrule(lr){4-5}\cmidrule(lr){6-7}
Method            &  R@5 & R@10 &  R@3 &  R@5 & R@50 & R@100 \\
\midrule
GPT-4 \cite{achiam2023gpt} (2D+depth)                          & 0.34 & 0.42 & 0.55 & 0.58 & 0.52 &  0.54 \\
Llama 3.2 \cite{dubey2024llama} (2D+depth)                          & 0.40 & 0.52 & 0.46 & 0.48 & 0.45 &  0.51 \\
\midrule
Open3DSG \cite{Koch_2024_CVPR}                           & 0.56 & 0.61 & 0.58 & 0.65 & 0.55 &  0.56 \\
$\text{ConceptGraphs}$  \cite{gu2024conceptgraphs}   & 0.37 & 0.46 & 0.74 & 0.79 & 0.69 &  0.71 \\
\textbf{\ours}                                            & \textbf{0.69} & \textbf{0.80} & \textbf{0.76} & \textbf{0.82} & \textbf{0.73} &  \textbf{0.74} \\
\bottomrule
\end{tabular}
\caption{\textbf{3D Scene Graph Prediction on 3DSSG.} \ours outperforms existing open-vocabulary 3D scene graph approaches as well as 2D-only frontier models. \ours can lift different frontier models into 3D with similarly strong performance.}
\label{tab:scenegraph_evaluation_rio}
\end{table}

\boldparagraph{Data}
In the following, we report quantitative 3D scene graph evaluation results on the RIO10 subset of the 3DSSG dataset \cite{Wald_2020_CVPR}. The 3DSSG dataset consists of semantic scene graphs for 3D point clouds and posed RGB-D frames obtained from a Google Tango device. It contains a closed vocabulary with 160 object classes and 27 relationship types.

\boldparagraph{Baselines} 
We compare our approach against ConceptGraphs \cite{gu2024conceptgraphs}, which also uses GPT-4, but in combination with a SLAM pipeline that predicts image captions. Once, the scene is reconstructed, GPT-4 is used to provide scene-consistent object and relationship caption. Additionally, we compare against Open3DSG \cite{Koch_2024_CVPR}, which uses a combination of CLIP \cite{pmlr-v139-radford21a}, and InstructBLIP \cite{instructblip} distilled into a 3D graph neural network. Furthermore, we propose additional 2D-based baselines for GPT-4 \cite{achiam2023gpt} and Llama 3.2 \cite{dubey2024llama}, which utilize recorded depth data to lift their 2D predictions to 3D.

\boldparagraph{Results}
A quantitative 3D scene graph comparisons is reported in \cref{tab:scenegraph_evaluation_rio}. 
We query the 160 object and 27 relationship classes and obtain the embedding similarity of the language feature with the feature field and treat the extracted similarity as a label confidence.
\ours demonstrates state-of-the-art results compared to other recent open-vocabulary 3D scene graph approaches and compared to ConceptGraphs \cite{gu2024conceptgraphs}. 
Our method demonstrates improved performance across all tasks: object, predicate as well as relationship prediction.
Furthermore, 2D methods exhibit inferior performance compared to the 3D approaches, potentially due to occlusions and view-dependent challenges. 
Please note, our approach, compared to closed-set segmentation methods does not require any semantic labels for training and can be deployed on any dataset that provides posed RGB frames.\looseness=-1

\begin{figure}
\vspace{0.25cm}
    \centering
    \includegraphics[width=0.90\linewidth]{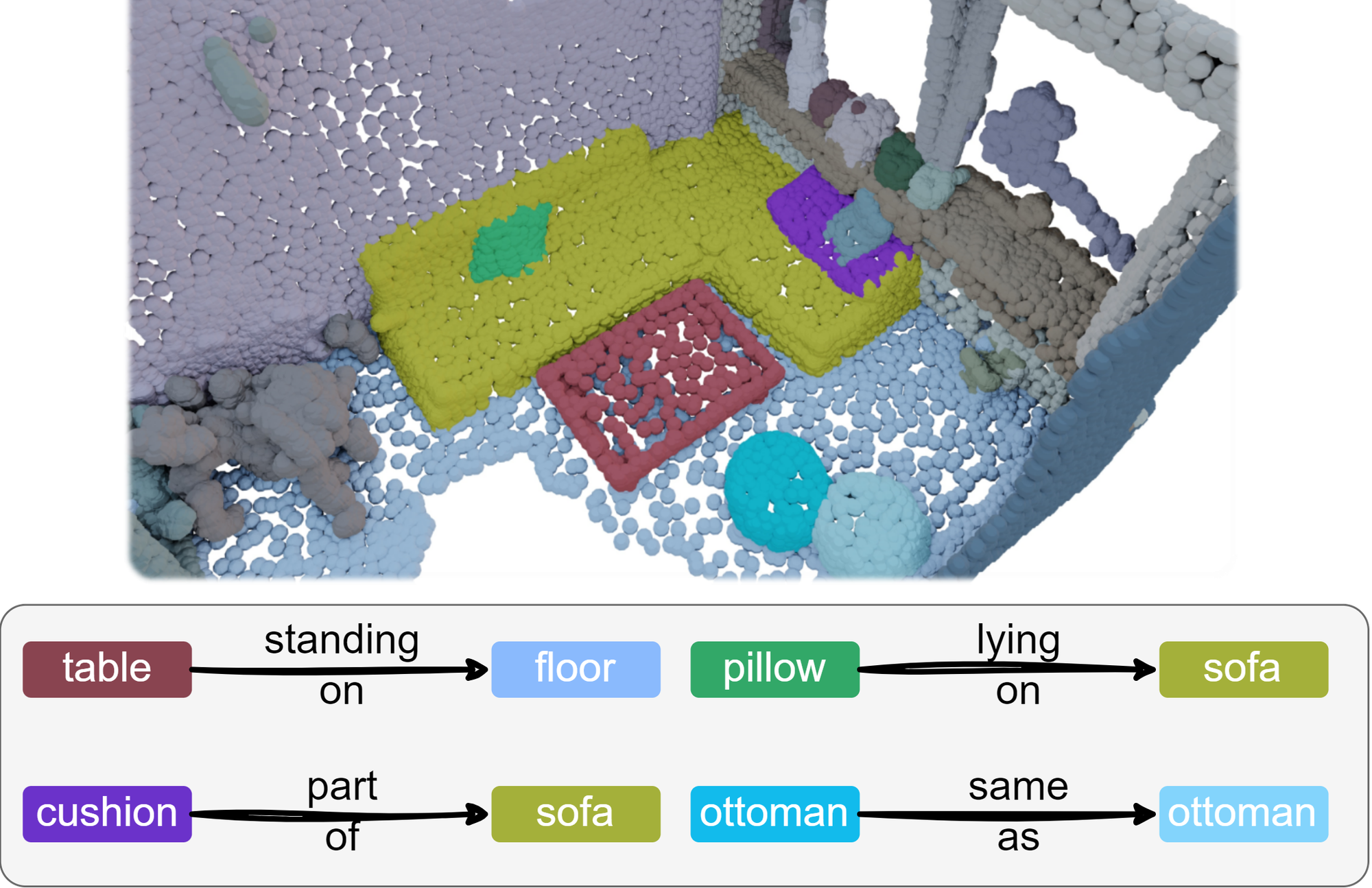}
    \caption{\textbf{3D Scene Graph Prediction.} Our open-vocabulary approach is able to predict complete 3D scene graph edges containing a subject-predicate-object relationship.}
    \label{fig:scene_graph}
\end{figure}

\cref{fig:scene_graph} show a subset of extracted relationships with subject, predicate, and object labels, respectively, on a scene from the 3DSSG dataset. For clarity, we omit the complete graph but show the most interesting relationships. More 3D scene graph results can be found in the supplementary.

\begin{figure}
    \centering
    \includegraphics[width=0.9\linewidth]{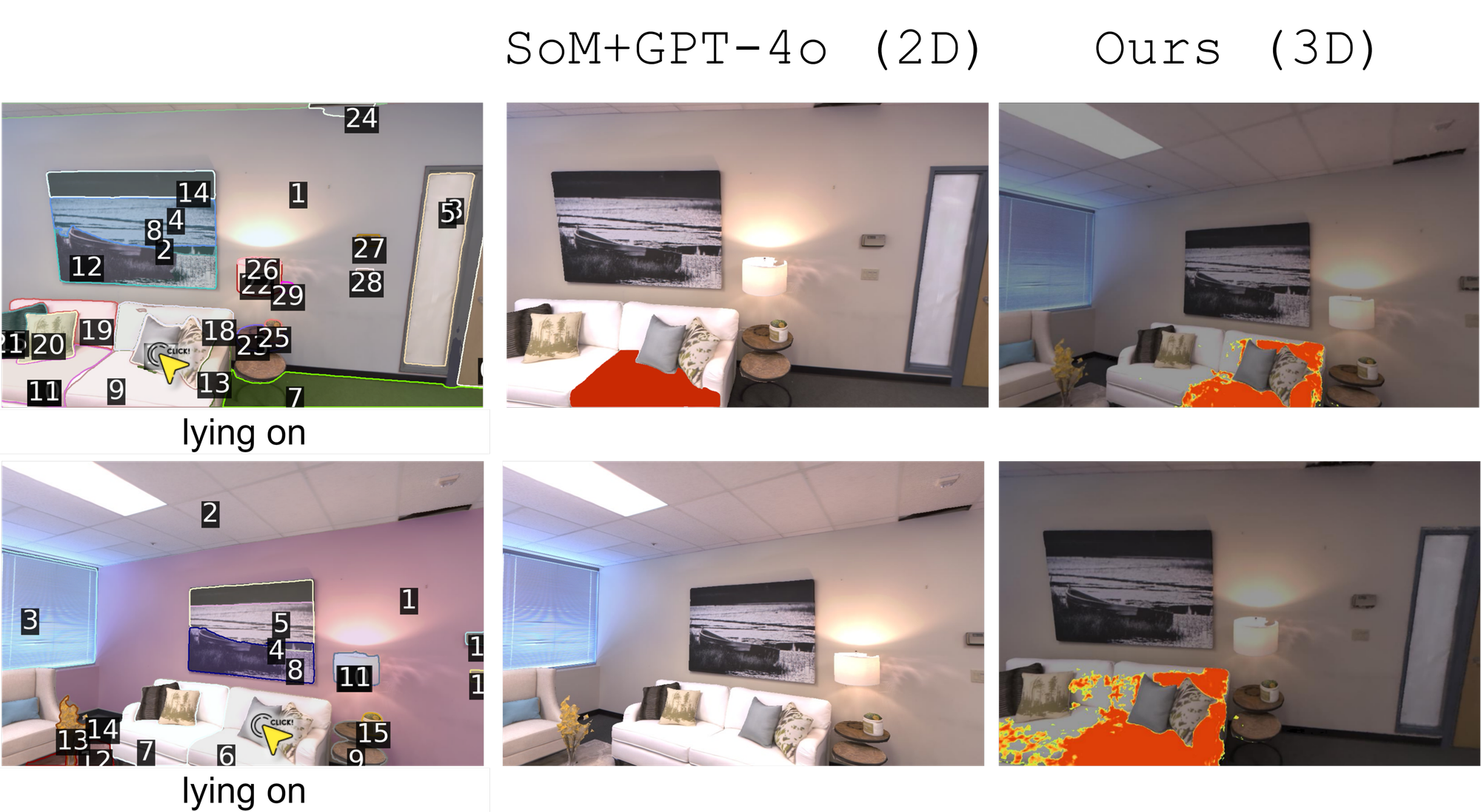}
    \caption{\textbf{3D Consistency Ablation.} 
    \textit{Left:} Extracted SoM marks per image with \includegraphics[scale=0.02]{fig/resources/click.png} query. \textit{Center:} Existing relationship in GPT-4 caption. \textit{Right:} Relationship response from \ours rendered into image space. While GPT-4 struggles with partially visible objects, \ours produces more robust results, independent of the view, because our volumetric rendering incorporates information from multiple views and models the underlying 3D relationship representation.
    }
    \label{fig:2d_vs_3d}
\end{figure}

\boldparagraph{Ablation -- Advantages of 3D relationship modeling over 2D inference} %
This paper demonstrates a process to distill knowledge from multi-modal LLMs such as \mbox{GPT-4} into a 3D consistent representation. In \cref{tab:scenegraph_evaluation_rio} and \cref{fig:2d_vs_3d}, we analyze the benefit of a 3D representation over a 2D-only approach which directly utilizes our knowledge provider \mbox{GPT-4}. It can be seen that the 2D approach will always suffer from view-dependent effects. \cref{fig:2d_vs_3d} shows how \mbox{GPT-4} is missing the \textit{lying on} relationship because some objects are only partially visible in the current frame. Meanwhile, when rendering the 3D prediction from \ours, our model is able to predict the correct relationships since it relies on the underlying 3D representation. The quantitative results confirm this observation, see~\cref{tab:scenegraph_evaluation_rio} where \ours clearly outperforms the 2D-only \mbox{GPT-4} model. This shows that our model generalizes beyond simple view-level supervision and, indeed, learns a consistent 3D representation, which improves over simple aggregated 2D inference.

\begin{figure}
    \centering
    \vspace{1em}
    \includegraphics[width=\linewidth]{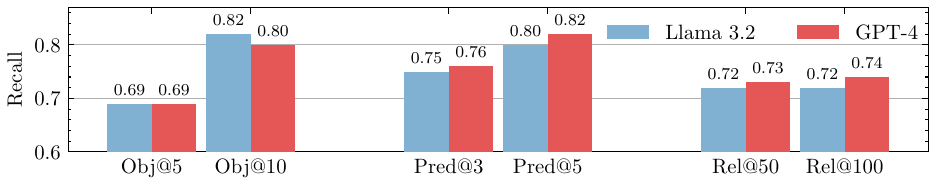}
    \caption{\textbf{Language Model Ablation.} We compare GPT-4 with Llama 3.2 as the relationship extractor of \ours for 3D scene graph prediction.}
    \label{fig:comparison}
\end{figure}

\noindent\textbf{Ablation -- Impact of multi-modal LLM choice on relationship understanding.}
While we utilize \mbox{GPT-4} as our backbone model for extracting relationships, our approach is agnostic to the backbone model and can accommodate any LLM capable of reasoning about object relationships. 
In \cref{fig:comparison}, we compare our approach which is using the latest version of \mbox{GPT-4o} against the popular open-source alternative Llama 3.2 \cite{dubey2024llama} (90B). \mbox{Llama 3.2}, which is considerably smaller than \mbox{GPT-4o}, has only a minor recall drop for relationship prediction. This shows that our model can be trained with any sufficiently powerful multi-modal LLM.

\subsection{Relationship-guided 3D Instance Segmentation} \label{sec:relationship_inst_refer}
To highlight the advantages of understanding relationships, we propose a new evaluation task for quantitative relationship-guided 3D instance segmentation. In this task, we want to highlight the benefit of understanding relationships from open-vocabulary textual descriptions for localizing objects of interest.

\boldparagraph{Data}
We label a small benchmark on Scannet++~\cite{Yeshwanth_2023_ICCV} of language-based relationship queries across 8 scenes with instance annotations for $\raisebox{0.3ex}{\texttildelow}30$ relationship queries spanning $\raisebox{0.3ex}{\texttildelow}40$ unique object types and $\raisebox{0.3ex}{\texttildelow}10$ semantic predicates. More details can be found in the supplementary.

\boldparagraph{Baselines}
For a fair comparison, we compare \ours against three state-of-the-art feature field methods for open-vocabulary object segmentation, LERF \cite{Kerr_2023_ICCV}, OpenNeRF~\cite{Engelmann_2024_ICLR}, and LangSplat~\cite{Qin_2024_CVPR} which all rely on posed RGB for training and inference. All approaches are able to process open-vocabulary queries in natural language and localize them in the 3D scene by associating the CLIP \cite{pmlr-v139-radford21a} embedding of the query with the learned features in the NeRF. 
The experiments show that our approach is the only capable method to reliably understand complex prompts such as ``the picture standing on the shelf" explicitly.

\boldparagraph{Localization}
To localize target queries with \ours, we split the language queries into nouns and verbs. First, the nouns are localized using the object field by computing the cosine-similarity to the nouns in the language query. Then, we refine the localization by combining the object prediction with the relationship embedding by rejecting all candidate predictions that do not have a relationship feature aligned with the verb from the query.
For LERF, OpenNeRF and LangSplat, the full query is processed directly, as these models do not distinguish between verbs and nouns in their query parsing.

\boldparagraph{Results}
In \cref{tab:scannetpp_instance_referral}, we report the segmentation accuracy and IoU for the set of target queries. The performance of LERF, OpenNeRF and LangSplat degrades in this specialized setting where all queries contain complex relationships. We observe most failure cases for duplicate objects where the \textit{bag-of-words} representation of CLIP cannot differentiate these objects by their relationship.
Meanwhile, \ours clearly outperforms LangSplat, OpenNeRF and LERF since it is able to model the relationship feature directly.

\begin{table}[t]
\tabcolsep=4.55mm
\centering
\scriptsize
\begin{tabular}{@{}l>{\raggedleft\arraybackslash}p{4cm}r@{}}
\toprule
Method                       & IoU & Acc  \\
\midrule
LERF \cite{Kerr_2023_ICCV}                              & 0.25   & 0.50 \\
OpenNeRF \cite{Engelmann_2024_ICLR}                     & 0.45 & 0.83 \\ 
LangSplat \cite{Qin_2024_CVPR} &  0.49 & 0.87 \\
\textbf{\ours}                         & \textbf{0.53}   & \textbf{0.96} \\
\bottomrule
\end{tabular}
\caption{\textbf{Open-Vocabulary relationship-guided Instance Segmentation.} Comparison of open-vocabulary radiance field-based methods on instance segmentation performance for challenging relationship queries.}
\label{tab:scannetpp_instance_referral}
\end{table}

\section{Limitations} \label{sec:limitations}
The experiments conducted in this paper demonstrate the potential and advantages of learning 3D relationships in radiance fields. However certain limitations remain. For instance, 
the relationship knowledge embedded in \ours is highly dependent on the multi-modal LLM prompting and its output.
Furthermore, while posed RGB recordings are easier to acquire than point clouds, \ours requires known calibrated camera intrinsics and
high-quality multi-view captures, which are not always
available or easy to capture. 
In general, the quality of \ours is bounded by the quality of the radiance field reconstruction.

\section{Conclusions} \label{sec:conclusion}

In this paper, we present \ours, the first 3D scene representation based on radiance fields that allow for open-vocabulary object and relationship queries. By distilling knowledge from 2D multi-modal LLMs into radiance fields, we are able to not only extract relationship information but also to obtain state-of-the-art open-vocabulary 3D scene graphs. 
We demonstrate that \ours effectively learns a consistent 3D representation that surpasses the performance of simple aggregated 2D inference.
Furthermore, we introduce a new task of relationship-guided 3D instance segmentation, to highlight the importance of understanding relationships for localizing objects of interest.
We hope this work will encourage future 3D scene understanding techniques to not only focus on object-centric features but explicitly incorporate the relations between them.

\vspace{0.5em}\noindent\textbf{Acknowledgement.}
We sincerely thank Jonathan Francis for providing support in running Llama 3.2 experiments. Our appreciation extends to David Adrian
for helpful discussions. We are also grateful to Timm Linder and Andrey Rudenko for their proofreading. This work was partly supported by the EU Horizon 2020 research and innovation program under grant agreement No. 101017274 (DARKO).

{
    \small
    \bibliographystyle{ieeenat_fullname}
    \bibliography{main}
}
\clearpage

\setcounter{page}{1}

\setcounter{section}{0}
\renewcommand\thesection{\Alph{section}}

\maketitlesupplementary

In this \textbf{supplementary material}, we first provide additional training details in \cref{sec:training_details}. Next, we offer further insights into our design choices for \ours in \cref{sec:design_choices}. \cref{sec:sg_construction} contains additional details on the scene graph extraction. We then present qualitative results for relationship segmentation and the relationship-guided 3D instance segmentation task in \cref{sec:more_quali}. In \cref{sec:gaussian_splat}, we include an adaption of \ours to Gaussian Splatting together with a quality comparison. Finally, we provide examples from our curated relationship-guided 3D instance segmentation benchmark in \cref{sec:task_imgs}.

\section{Training details} \label{sec:training_details}
To accelerate training speed, existing feature field approaches \cite{Kerr_2023_ICCV, Kim_2024_CVPR, Engelmann_2024_ICLR} store extracted 2D training features to disk and load them in RAM at training start for efficient retrieval at each training step instead of computing the features online. 
However, storing all relationship features in RAM or disk is infeasible for \ours since for $n$ input images of shape $w,h$, with $m$ generated masks it would require storing $n \times m \times (m-1) \times w\times h\times d$ relationship features. This would result in $\raisebox{0.3ex}{\texttildelow}5.66TB$ when we store the features in FP16 and assume 10 instances per image for a scene of 200 images each with VGA resolution of $640\times 480$.
Instead, we optimize the memory resources by storing a dictionary of all relationship features in combination with a singular segmentation mask and compute the relationship map for each sampled pixel-pair using a two-step lookup in the segmentation map and then in the relationship dictionary. Using this strategy we are able to reduce the memory requirements to $\raisebox{0.3ex}{\texttildelow}500MB$ per scene when using FP16 precision.
Inspired by \cite{Kim_2024_CVPR}, we begin training the relationship field after 2000 steps of NeRF optimization to let the geometry converge. 
We train for 30000 steps on a single \mbox{Nvidia A100}, which takes around 60 minutes and consumes around 40GB of GPU memory. 
The feature extraction of the object features from OpenSeg \cite{OpenSeg_2022_ECCV} and SAM \cite{Kirillov_2023_ICCV}, as well as the relationship features with \mbox{GPT-4} 
increases the training time by about 30 minutes for the first run.

\section{Design choices}
\label{sec:design_choices}
\boldparagraph{Prompting}
To extract textual relationships using GPT-4~\cite{achiam2023gpt} or Llama \cite{dubey2024llama}, we employ a combination of visual and textual prompting. For visual prompting, we utilize SoM \cite{yang2023setofmark} to overlay semi-transparent masks and numeric marks. The textual prompt consists of a two-stage approach which queries the model first to extract objects by their mark-id and then to extract relationships referenced by the previously extracted object-ids together with a relationship label. The complete prompt looks as follows:
\begin{tcolorbox}[colback=black!5!white, colframe=black!75!black, left=2pt, right=2pt, top=2pt, bottom=2pt]
\footnotesize
\textbf{1. Object Identification:} Identify all objects in the image by their tag. Create a dict that maps tag\_id to class\_name.

\vspace{1em}
\textbf{2. Affordance/Relationship Detection:} For every pair of tagged objects that are clearly related, describe the semantic relationships and affordances as a list of dictionaries using the format [s\_id: \#n1, subject\_class: x, o\_id: \#n2, object\_class: y, predicates: [p1, p2, ...]]. For subjects and objects sharing multiple relationships/affordances, concatenate predicates with a comma in the [predicate] field.

\vspace{1em}
- Avoid generic terms like "next to" for ambiguous relationships. Instead, specify relationships with precise relationships and affordances describing spatial relationships [over/under etc.], comparative relationships [larger/smaller than, similar/same type/color], functional relationships [part of/belonging to, turns on], support relationships [standing on, hanging on, lying on, attached to].

- Do not use left/right; always use 3D consistent relationships.

- Always combine a spatial relationship with a semantic, comparative, functional or support relationship using a comma (e.g., [A] [above, lying on] [B]).

- For symmetrical relationships, include both directions (e.g., [A] [above] [B] and [B] [below] [A]).

- Even for distant objects highlight if they are [same/similar/same color/same object type]

\vspace{1em}
\textbf{Example Output:}

objects = \{4: floor, 7: table, 12: chair, ...\}

relationships\_affordances = \{

    [s\_id: 4, subject\_class: table, o\_id: 7, object\_class: floor, predicates: standing on],
    
    [s\_id: 12, subject\_class: chair, o\_id: 13, object\_class: chair, predicates: next to, same as],
    
    [s\_id: 6, subject\_class: pillow, o\_id: 8, object\_class: couch, predicates: belongs to],
    
    ...
    
\}
\end{tcolorbox}
\noindent After processing the image frames with the LLM we parse the output into a JSON format in an automatic manner.

\boldparagraph{Text encoder}
\begin{figure*}
    \centering
    \includegraphics[width=0.95\linewidth]{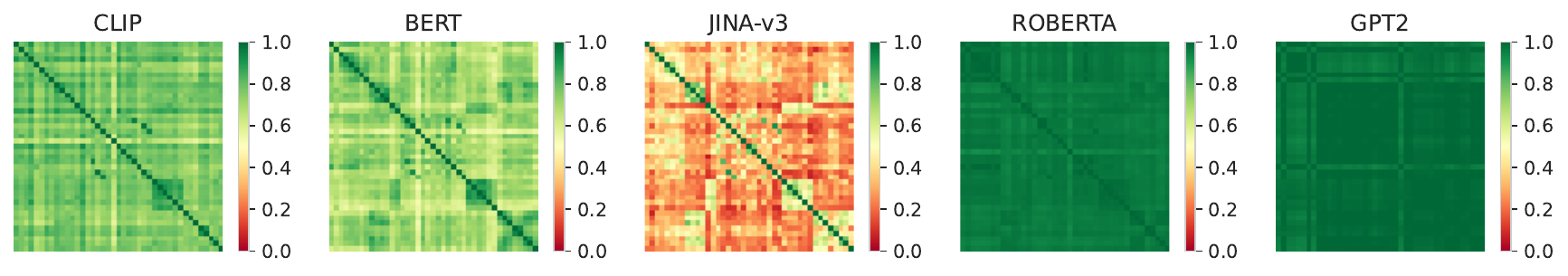}
    \caption{\textbf{Language Encoder Ablation.} We compare 5 language encoder-only model based on their separability in embedding space. For each language encoder, we plot a similarity matrix, for the pairwise cosine-similarity of 41 predicates taken from the 3DSSG dataset.
    }
    \label{fig:similarity_matrix}
\end{figure*}

To embed relationships in \ours, we encode the output from a multi-modal LLM into the radiance field using an encoder-only language model. The choice of the encoder is important since it determines the structure and queryablity of the embedding space in the radiance field. We want an embedding space, that is highly structured and embeds similar (relationship) concepts close together, while contradictory relationships are supposed to be far apart in embedding space. 
In \cref{fig:similarity_matrix}, we provide an analysis for different popular open-source text encoders such as CLIP \cite{pmlr-v139-radford21a}, BERT \cite{kenton2019bert}, Jina-v3 \cite{jina-v3}, RoBERTa \cite{roberta} and GPT-2 \cite{radford2019language}. We have a set of 41 distinct relationships with varying semantic similarity to each other and plot their pair-wise cosine similarity in a similarity matrix. 
We observe that Jina-v3-embeddings generate the most well-structured feature space, where related concepts exhibit a strong similarity, while the majority of relationships describing distinct concepts show a high degree of dissimilarity.
As a counter-example, both RoBERTa and GPT2 embed all relationships in a very similar feature space, which would make fine-grained querying difficult.

\begin{figure}
    \centering
    \includegraphics[width=0.9\linewidth]{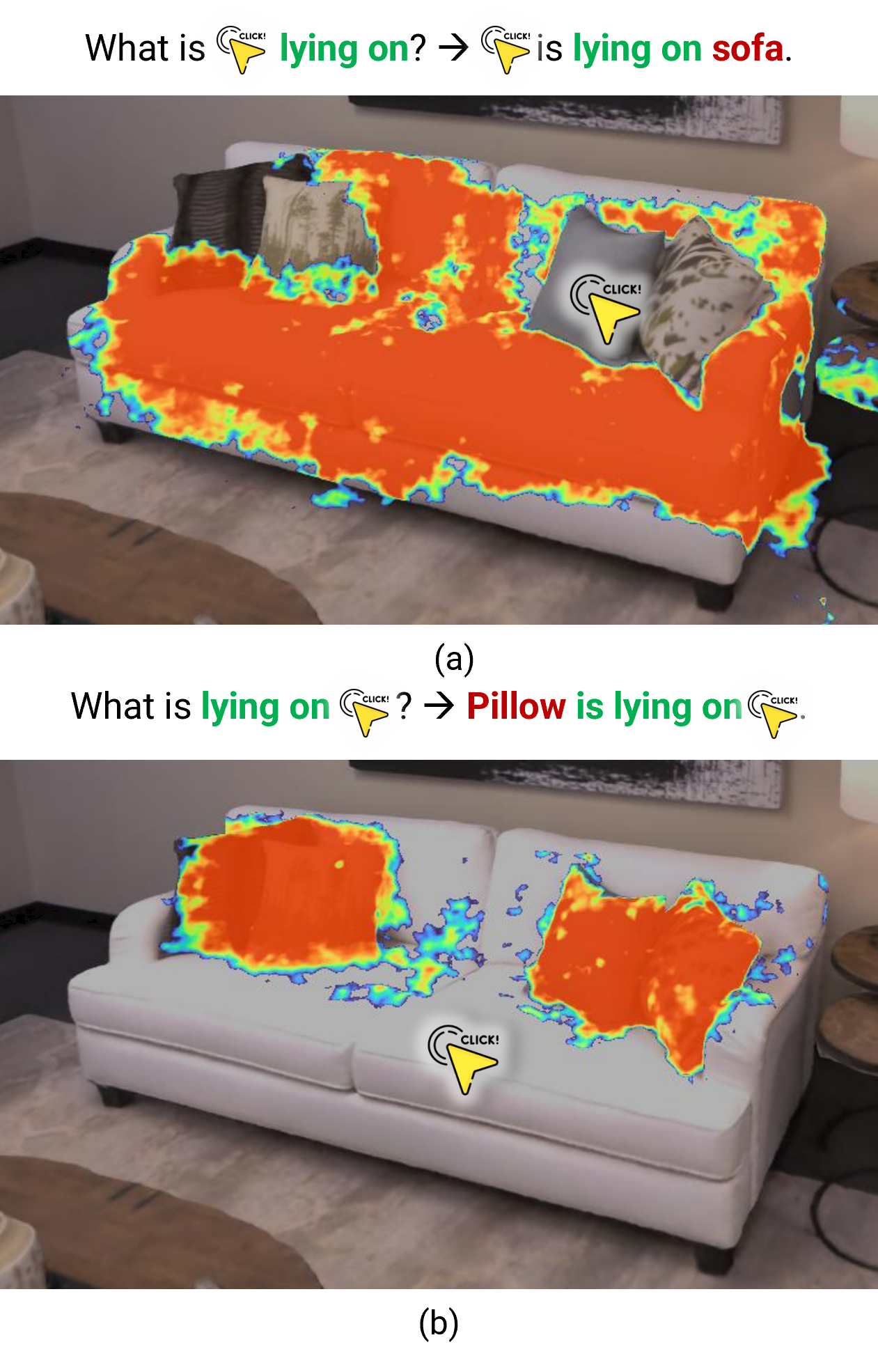}
    \caption{\textbf{Relationship Direction.} (a) visualizes the relationship response for the question ``What is \includegraphics[scale=0.025]{fig/resources/click.png} \textit{standing on/lying on/similar to}?", where we localize the object in a \textit{subject-predicate-object} relationship. While (b) visualizes the relationship response for the question ``What is \textit{standing on/lying on/similar to} \includegraphics[scale=0.025]{fig/resources/click.png}?", where we localize the subject in a \textit{subject-predicate-object} relationship.
    }
    \label{fig:direction}
\end{figure}

\boldparagraph{Relationship direction}
In Fig. 3 of the main manuscript, we present qualitative results from \ours on 4 scenes. In these results, we present queries of the form ``What is $<$\includegraphics[scale=0.025]{fig/resources/click.png}$>$ \textit{standing on/attached to/similar to etc.}?". In this scenario, we are interested in the object of a \textit{subject-predicate-object} relationship. 
However similarly, it can be interesting to investigate to query the subject of a \textit{subject-predicate-object} relationship by answering the question ``What is \textit{standing on/attached to/similar to etc.} $<$\includegraphics[scale=0.025]{fig/resources/click.png}$>$?" To model this question in \ours we simply have to invert the supervision signal during training by swapping the query ray origin with the ray origin.
In \cref{fig:direction}, we demonstrate different directional relationship queries for the same objects and predicate.

\section{Scene Graph Construction} \label{sec:sg_construction}
\begin{figure*}
    \centering
    \includegraphics[width=0.95\linewidth]{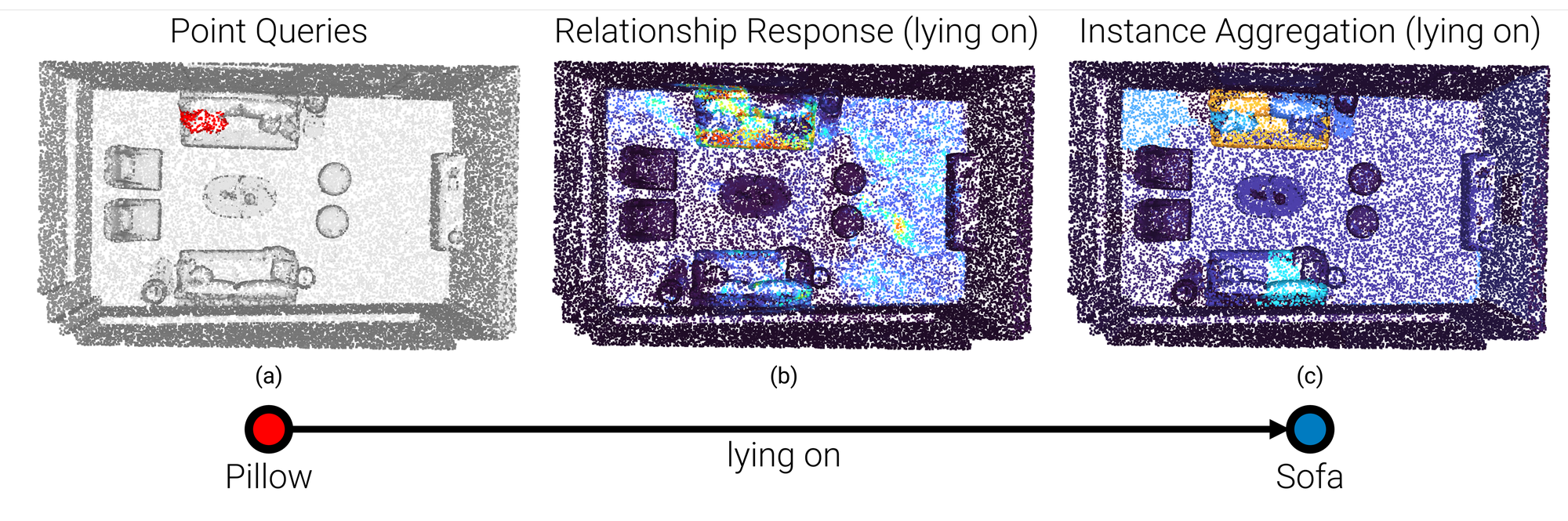}
    \caption{\textbf{Relationship Edge Construction.} To extract a 3D scene graph from \ours, we automatically query instances (a), compute the relationship response for predicates such as ``lying on" (b), and aggregate the relationship response for each instance (c). We add an edge to the scene graph for all objects whose relationship response for the subject and predicate is greater than a certain threshold.
    }
    \label{fig:scene_graph_construction}
    \vspace{1ex}
\end{figure*}

In \cref{fig:scene_graph_construction}, we visually supplement the reported process of extracting a 3D scene graph from \ours. First, we extract groups of points from the instance field. These groups of points (\cref{fig:scene_graph_construction}a), serve as the queries for the \ours and represent the subject in a \textit{subject-predicate-object} relationship edge. In a second step, the \ours gets evaluated on the remaining points of the point cloud given the query points and a textual relationship prompt such as lying on (\cref{fig:scene_graph_construction}b). The textual query represents the predicate in the \textit{subject-predicate-object} relationship. This step returns a relationship activation map for the entire point cloud with each point having a unique relationship response. In the third step, the activations get aggregated based on the instance head (\cref{fig:scene_graph_construction}c). The instances that have a relationship response greater than a threshold of 0.5 represent objects in the \textit{subject-predicate-object} relationship edge. Finally all edges for objects surpassing the threshold are added to the 3D scene graph.
\todo[inline]{conceptgraphs comparison notes}

\begin{figure*}
    \centering
    \includegraphics[width=\linewidth]{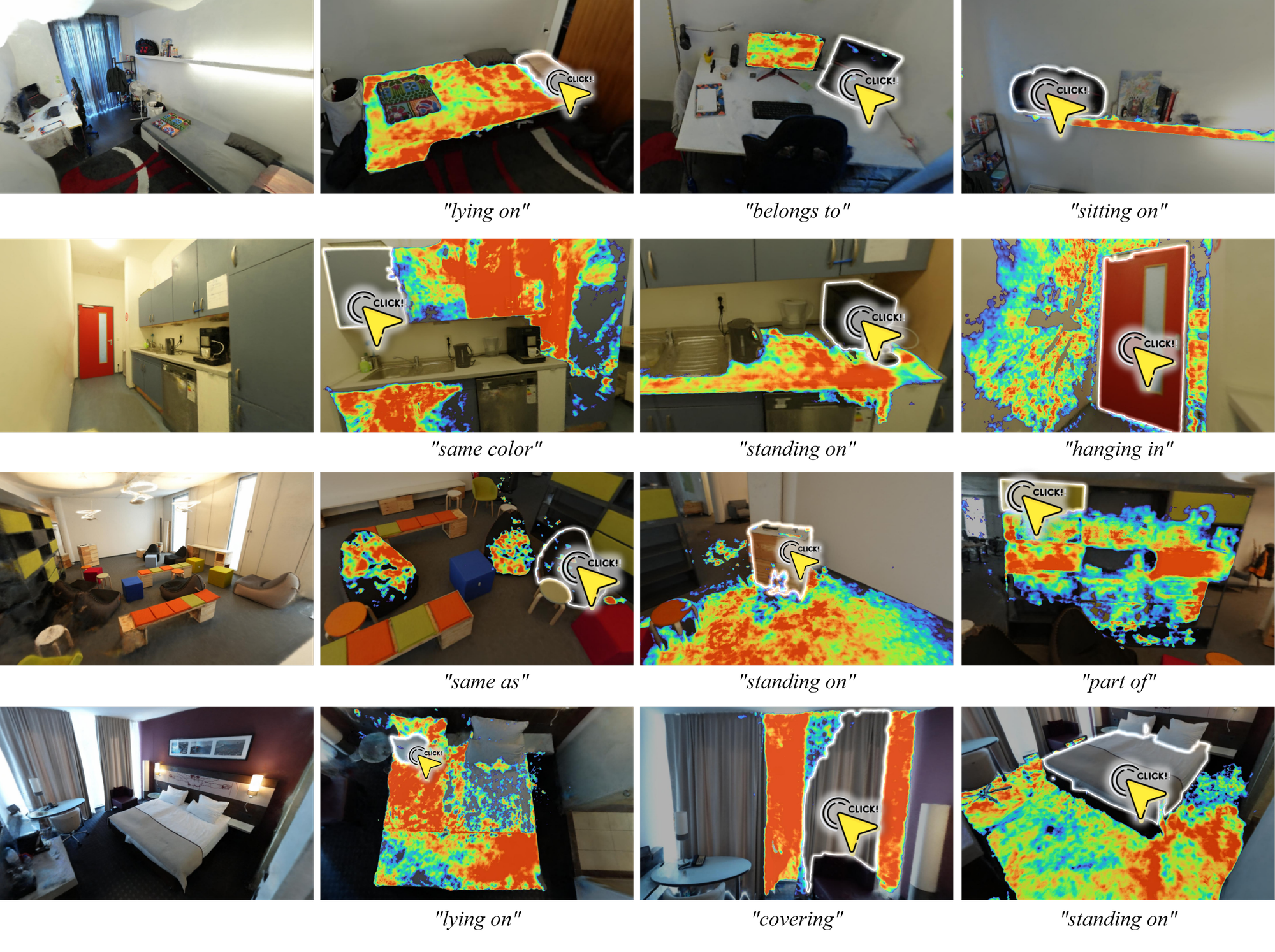}
    \vspace*{-20pt}
    \caption{\textbf{Additional Qualitative Results.} We provide relationship responses for 4 additional scenes from Scannet++. The colormap visualizes the relationship response where blue is low and red is high. We visualize the relationships for the question: ``What is \includegraphics[scale=0.025]{fig/resources/click.png} \textit{standing on/lying on/similar to}?"
    }
    \label{fig:more_examples}
    
\end{figure*}

\begin{figure*}[t]
    \centering
    \includegraphics[width=0.75\linewidth]{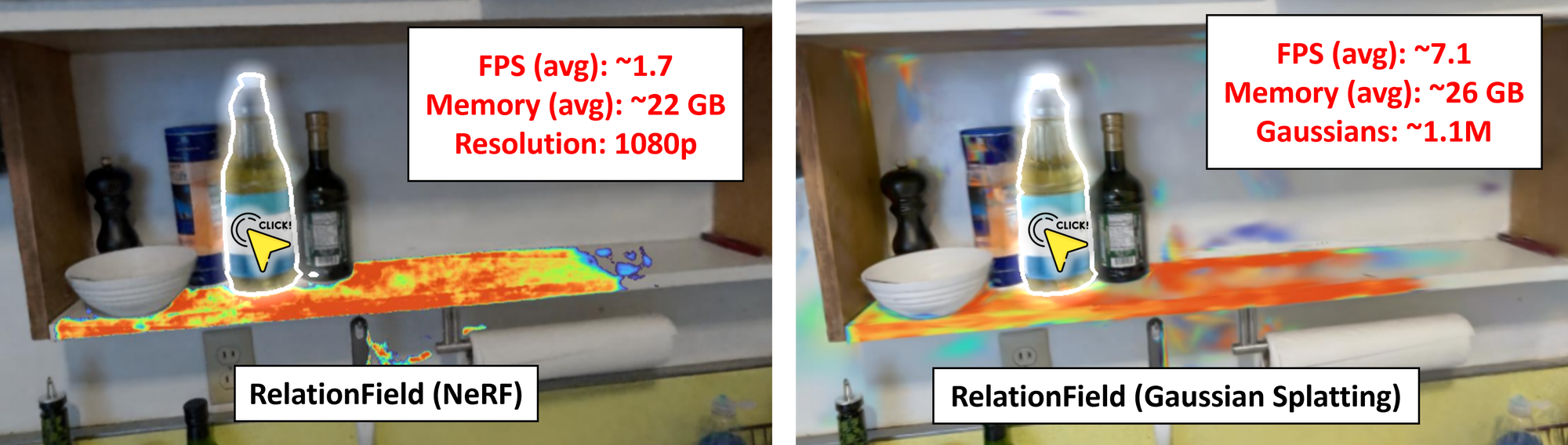}
    \caption{\textbf{RelationField w/ NeRF or w/ Gaussian Splatting geometry}. We compare the rendering speed (FPS), memory requirements and RelationField quality for the query ``standing on".}
    \label{fig:gaussian_vs_nerf}
\end{figure*}

\section{Qualitative Results} \label{sec:more_quali}
\boldparagraph{Relationship querying}
In \cref{fig:more_examples}, we present qualitative results for 4 additional scenes for the relationship querying with \ours.

\begin{figure*}
    \centering
    \includegraphics[width=0.95\linewidth]{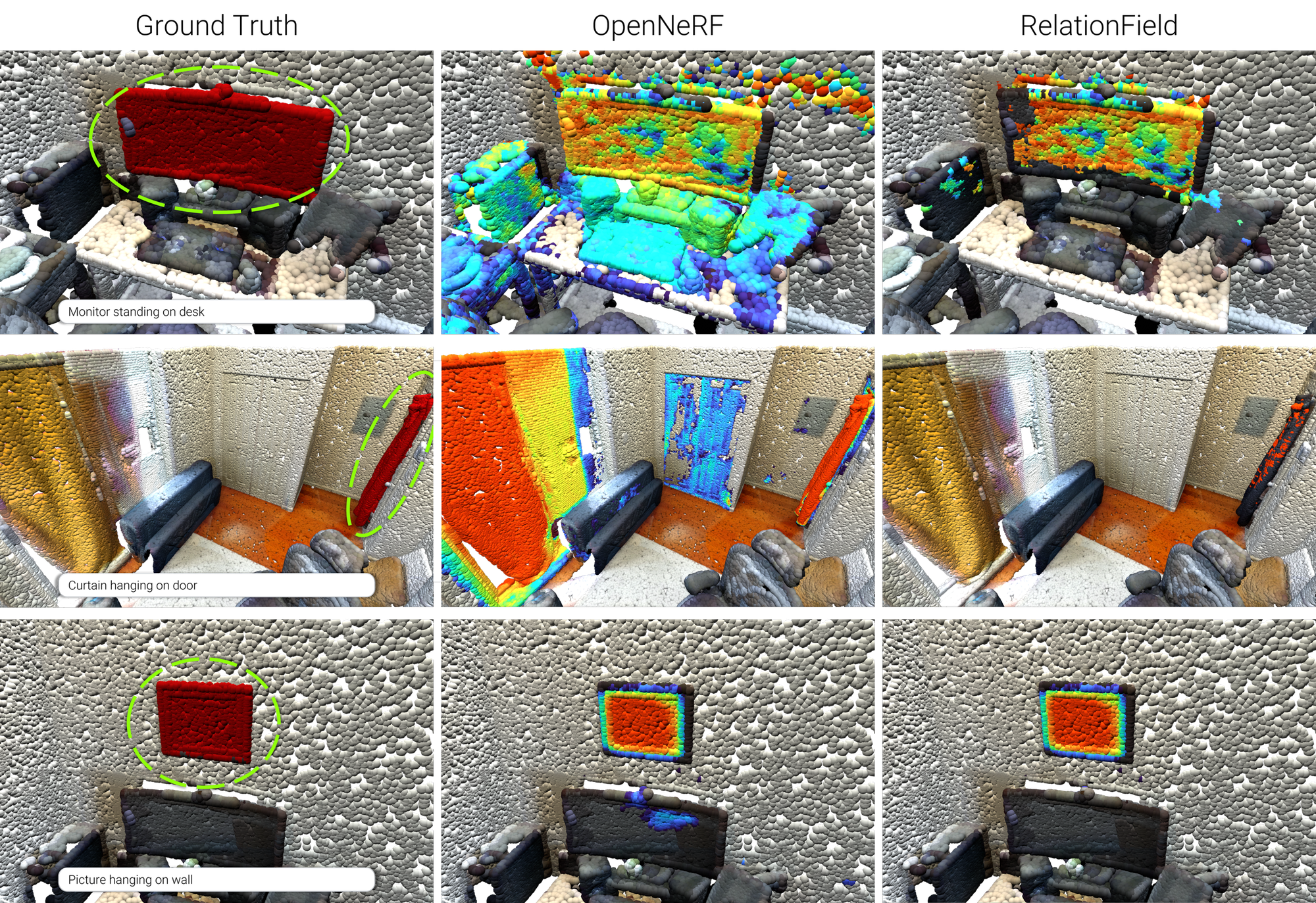}
    \caption{\textbf{Releationship-guided 3D Instance Segmentation.} We compare OpenNeRF with \ours for relationship-guided 3D instance segmentation. While OpenNeRF produces many false positives because it gets confused with compositional queries arising from the \textit{bag-of-words} behavior of CLIP. Meanwhile \ours uses the object information together with the relationship information from the prompt to accurately filter predictions that only correspond to the subject in the prompt.
    }
    \label{fig:relseg_compare}
\end{figure*}

\boldparagraph{Relationship-guided instance segmentation} In \cref{fig:relseg_compare}, we qualitatively compare the 3D instance segmentation of \ours against OpenNeRF \cite{Engelmann_2024_ICLR} for relationship queries to supplement Tab. 2. OpenNeRF produces many false positives because it gets confused with the compositional queries arising from the bag-of-words behavior of CLIP \cite{radford2019language}. Meanwhile, RelationField uses the object information together with the relationship information from the prompt to accurately filter predictions that only correspond to the object in the prompt that has the described relationship.
\todo[inline]{supplementary video?}

\section{Gaussian Splatting Support}\label{sec:gaussian_splat}
We build \ours on NeRF \cite{Mildenhall_2020_ECCV}, however since our approach is independent of the underlying 3D representation, it is possible to extend \ours to Gaussian Splatting for faster training, inference and rendering. To train \ours, we follow \cite{lee2024open3drf} and initialize the Gaussian Splatting training run with the exported point cloud of the NeRF training. This results in faster convergence and fewer Gaussians leading to improved memory utilization. 
For \ours with Gaussian Splatting geometry, we reformulate our relationship definition from a pair of rays to a pair of 3D Gaussian centers. In \cref{fig:gaussian_vs_nerf}, we compare the rendering speed, memory requirements and RelationField quality. \ours based on Gaussian Splatting achieves 4x faster rendering compared to its NeRF variant with a lower memory footprint. Overall, \cref{fig:gaussian_vs_nerf} shows that \ours is independent of the underlying geometry, and both NeRF and 3DGS produce high-quality RelationFields.

\section{Relationship-guided 3D Instance Segmentation Dataset} \label{sec:task_imgs}
\begin{figure*}
    \centering
    \includegraphics[width=0.95\linewidth]{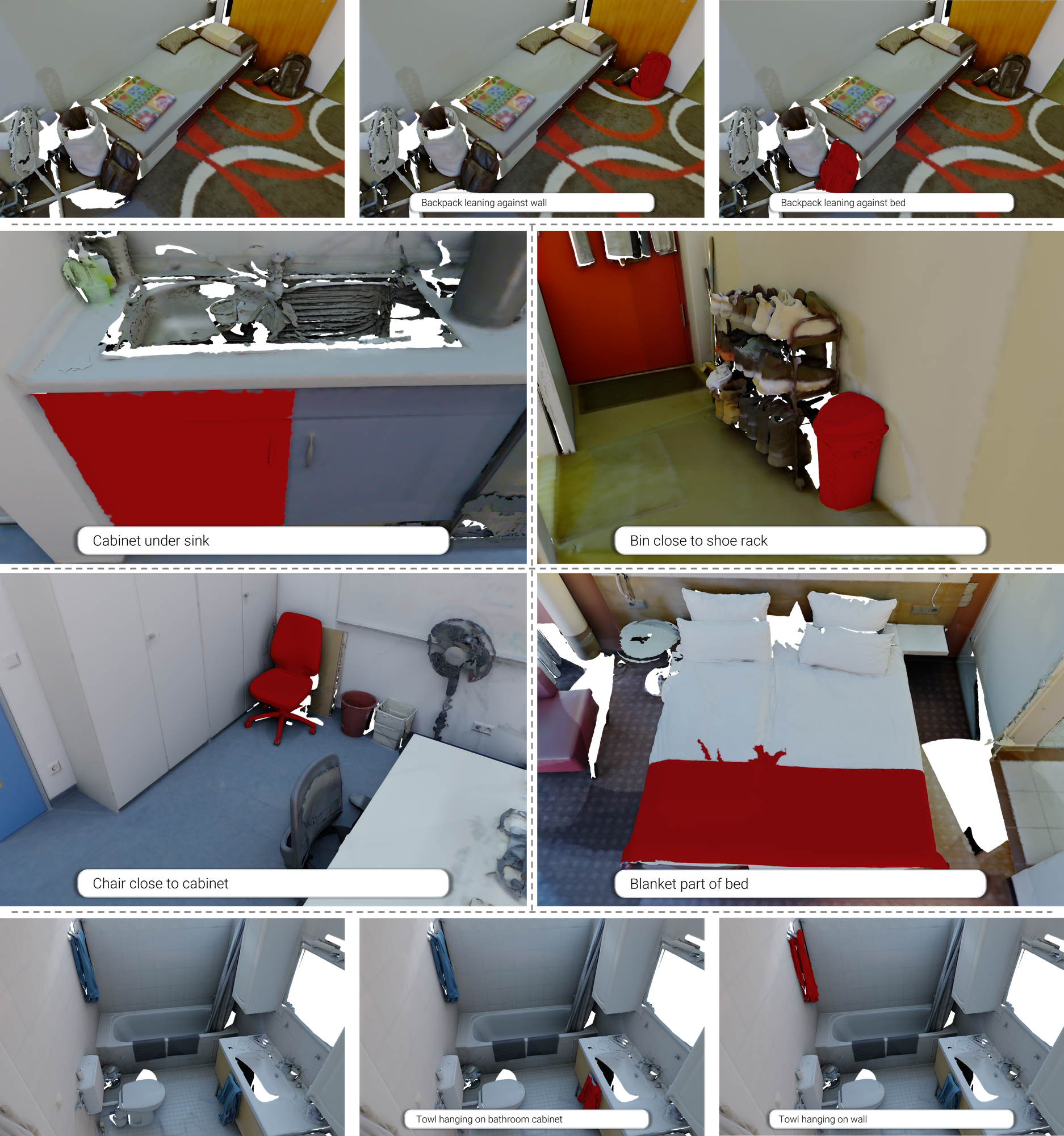}
    \caption{\textbf{Releation-guided 3D Instance Segmentation Task Overview.} We visualize a few annotated segments from our labeled benchmark on Scannet++ together with annotated relationship prompts. We focus on objects which appear multiple times in the scene, but that can be uniquely referenced by a relationship prompt.
    }
    \label{fig:task_overview}
\end{figure*}

In \cref{fig:task_overview}, we present a subset of the annotated benchmark which we present in Sec 4.1 of the main paper. In the benchmark, we provide instance segmentations paired with textual relationship prompts. When curating the benchmark we focused on samples that appear multiple times in the scene, but which can be uniquely referenced by a relationship prompt.

\end{document}